\documentclass[10pt,twocolumn,letterpaper]{article}

\usepackage{cvpr}              

\usepackage[dvipsnames]{xcolor}

\definecolor{cvprblue}{rgb}{0.21,0.49,0.74}
\usepackage[pagebackref,breaklinks,colorlinks,citecolor=cvprblue]{hyperref}
\usepackage{multirow}
\usepackage{float}
\usepackage{pgfplots}

\def\paperID{7401} 
\def\confName{CVPR}
\def\confYear{2024}

\title{\LARGE \bf
UniDoorManip: Learning Universal Door Manipulation Policy Over Large-scale and Diverse Door Manipulation Environments
}

\author{
Yu Li\footnotemark[1] \textsuperscript{1} \quad 
Xiaojie Zhang\footnotemark[1] \textsuperscript{ 1} \quad
Ruihai Wu\footnotemark[1] \textsuperscript{ 2} \\
Zilong Zhang \textsuperscript{1} \quad 
Yiran Geng \textsuperscript{2} \quad 
Hao Dong\footnotemark[2] \textsuperscript{2} \quad
Zhaofeng He\footnotemark[2] \textsuperscript{ 1} \\
\textsuperscript{1}Beijing University of Posts and Telecommunications \quad
\textsuperscript{2}School of CS, Peking University \\
}

\begin{document}

\twocolumn[{
\renewcommand\twocolumn[1][]{#1}
\maketitle

\begin{center}
    \centering
    \captionsetup{type=figure}
    \includegraphics[width=1 \linewidth]{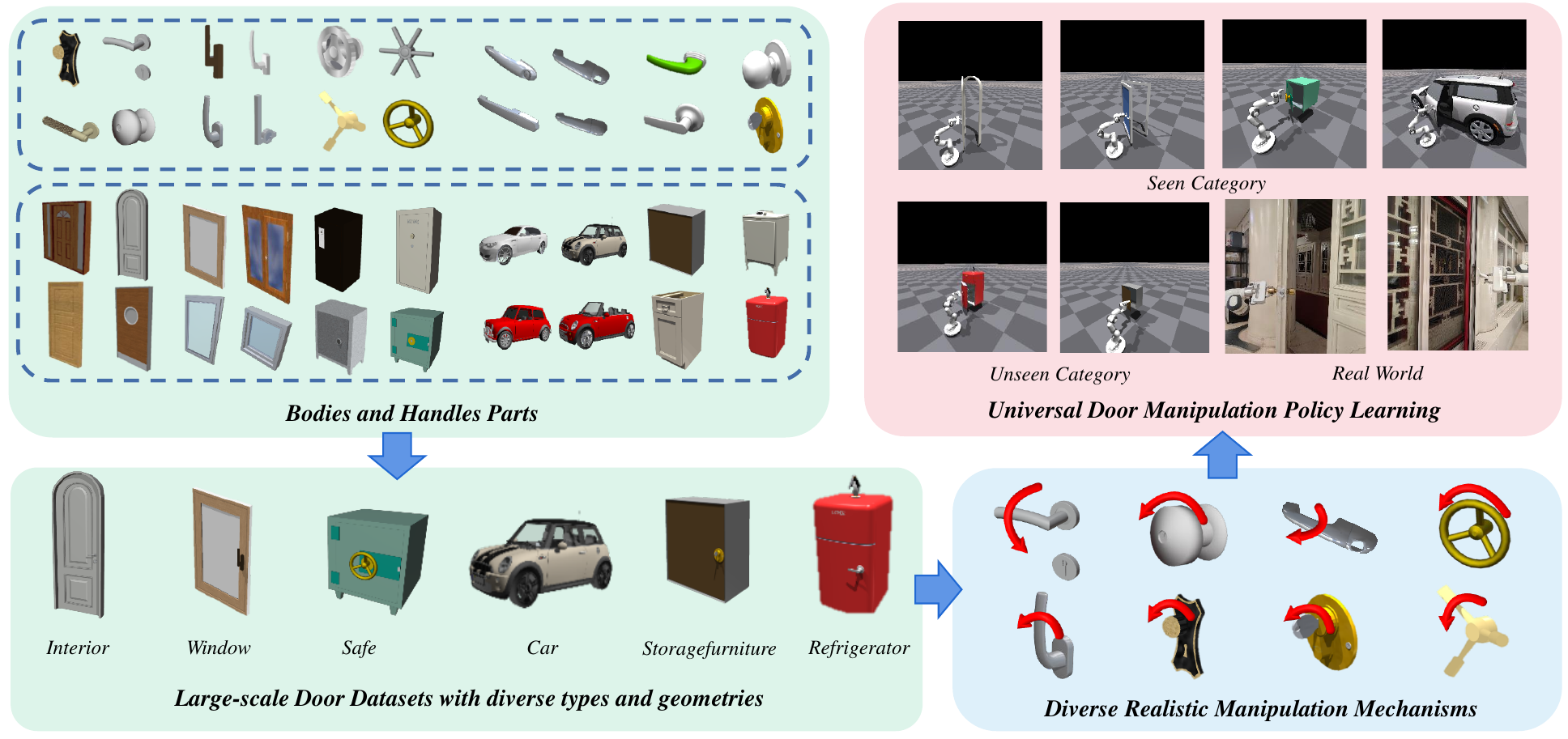}
    \caption{
    \textbf{Our Proposed Environment, Dataset and Universal Manipulation Policy.} We build a novel door manipulation environment equipped with a large-scale door dataset covering 6 door categories with hundreds of door bodies and handles, and configure it with different realistic door manipulation mechanisms. To learn the universal door manipulation policy, we propose a novel framework which can generalize to unseen shapes and categories. }
    \label{fig:teasor}
\end{center}
}]
\renewcommand{\thefootnote}{\fnsymbol{footnote}}
\footnotetext[1]{Equal contribution.}
\footnotetext[2]{Corresponding author.}

\begin{abstract}
Learning a universal manipulation policy encompassing doors with diverse categories, geometries and mechanisms, is crucial for future embodied agents to effectively work in complex and broad real-world scenarios. 
Due to the limited datasets and unrealistic simulation environments, previous works fail to achieve good performance across various doors.
In this work, we build a novel door manipulation environment reflecting different realistic door manipulation mechanisms, 
and further equip this environment with a large-scale door dataset covering 6 door categories with hundreds of door bodies and handles,
making up thousands of different door instances.
Additionally,
to better emulate real-world scenarios, we introduce a mobile robot as the agent 
and use the partial and occluded point cloud as the observation,
which are not considered in previous works while possessing significance for real-world implementations.
To learn a universal policy over diverse doors, we propose a novel framework disentangling the whole manipulation process into three stages,
and integrating them by training in the reversed order of inference.
Extensive experiments validate the effectiveness of our designs and demonstrate our framework's strong performance. Code, data and videos are avaible on \href{https://unidoormanip.github.io/}{https://unidoormanip.github.io/}.
\end{abstract}
\vspace{-4mm}

\section{INTRODUCTION}

 To enable robots to exhibit human-like abilities in performing a wide range of tasks, 
 it is crucial for them to acquire proficiency in manipulating articulated objects. Among these tasks, door manipulation holds significant importance due to the frequent need to open or close doors in various scenarios. 
 While previous works 
 have focused primarily on interior doors~\cite{urakami2019doorgym, wang2020learning}, we aim to extend doors to a more general setting, \emph{e.g.}, doors in windows, cars, safes, as illustrated in Figure~\ref{fig:teasor}.
 In the above broad scenarios, the door manipulation task covers doors with diverse types, geometries and manipulation mechanisms, which poses a great challenge to learn a universal door manipulation policy.

 Prior works in the field have struggled to learn the universal manipulation policy due to the lack of diversity in terms of types, geometries, and manipulation mechanisms. 
 Besides using push or pull to open the door\cite{Mo_2021_ICCV, xu2022umpnet, wu2022vatmart, geng2023gapartnet, geng2023partmanip, eisner2022flowbot3d},
 DoorGym~\cite{urakami2019doorgym} presents an approach that automatically generates door bodies and handles with hard-code, which results in bad performance when faced with new unseen doors due to the limited geometric diversity. To facilitate the learning of a universal door manipulation policy, we build a comprehensive door manipulation environment that encompasses doors across diverse types, geometries, and manipulation mechanisms. Our efforts in constructing this environment have been focused on the following two key aspects.

 Firstly, recognizing the limited door types, geometries and quantities in existing datasets~\cite{geng2023gapartnet, liu2022akb,  xiang2020sapien, urakami2019doorgym}, we propose a large-scale door dataset with diverse categories and geometries. Our dataset consists of two door components: body and handle, providing users the flexibility to configure doors according to their specific requirements, as illustrated in Figure~\ref{fig:teasor}. 
 Table~\ref{tab:datasets} shows that our dataset encompasses 6 distinct types of doors, comprising 328 door bodies and 204 handles, enabling the composition of thousands of door objects while ensuring their compatibility.  Secondly, to mitigate the gap between simulation and the real world, we introduce more realistic settings in our door manipulation environment. Concretely, our goal is to address the intricacies associated with doors featuring latching mechanisms, which require handle grasping and manipulation prior to the door opening. Compared with previous works utilizing sole parallel gripper or suction gripper~\cite{Mo_2021_ICCV, wu2022vatmart, xu2022umpnet, eisner2022flowbot3d}, we employ a mobile robot arm equipped with a parallel gripper as our agent. 
Furthermore, due to occlusion between the door and the robot arm, as well as occlusion within the door itself, our visual observations during the door manipulation process are partially occluded.


 To manipulate doors with latching mechanisms, previous works~\cite{urakami2019doorgym, geng2023partmanip} have explored the training of a single universal policy using reinforcement learning (RL) algorithms. 
 However, directly training such a policy for the entire door manipulation process in an end-to-end manner poses great challenges. 
 This is because door manipulation contains three separate but related stages: handle grasping, handle manipulation, and door opening. The inherent separation of these stages results in complicated manipulation mechanisms and vast exploration space, making it difficult for a single RL policy to learn. 
 To tackle this challenge, we propose a novel framework that disentangles these three stages, each stage contains a specific manipulation process, making it easier to learn a generalizable manipulation policy. Besides, we train these policies in a conditional way to reveal their interrelations and dependencies.
 
 In the three-stages manipulation, the first stage, handle grasping, only requires the grasp pose of the end-effector, while the latter two stages require action sequence generation. Hence we employ a policy specifically for predicting handle grasping action. 
 Leveraging the inherent generalizability across diverse geometries provided by visual affordance~\cite{gibson1977theory, wu2023learningforesightful}, the policy predicts a point-level score map where a higher score indicates a greater chance of successful door manipulation, to propose the grasp action. 
 After grasping the handle steadily, the goal of the following stages is to manipulate the handle to unlock the door and open the door. Unlike most works configuring the door with a simple manipulation mechanism, our proposed environment simulates diverse realistic door manipulation mechanisms that closely resemble real-world scenarios, \emph{e.g.}, lever, round, key and valve, as shown in the right part of Figure~\ref{fig:teasor}.
 Although handle manipulation and door opening share the similarity of generating action sequences, their specific action types differ significantly. 
 Hence we train separate policies for each stage to alleviate the burden of the model, enabling a universal capability for handle manipulation and door opening. In contrast to open-loop manipulation~\cite{wu2022vatmart}, which lacks real-time adjustments, we employ a closed-loop formulation that dynamically adapts subsequent actions based on the current observation for these two stages. Furthermore, to seamlessly integrate the three separate but related universal policies for each stage into a single universal policy for the entire manipulation process, we introduce a conditioned training strategy that bridges the gap among policies.
 
 Through extensive experiments in simulation, we validate the effectiveness of our design choices and demonstrate that our approach significantly outperforms previous methods. Additionally, we conduct real-world experiments to show the generalization capability of our approach in real-world scenarios.
 
 In summary, our main contribution encompasses:
 \begin{itemize} 
    \item We are the first to build a door manipulation environment with diverse realistic manipulation mechanisms and equip this environment with a large-scale door dataset covering diverse types, handles, and geometries for universal manipulation policy learning. 
    \item To achieve universal and realistic door manipulation, we propose a novel framework that disentangles the whole manipulation process into three stages with respective universal policies and integrates them into the whole universal policy leveraging conditioned training.
     \item Extensive experiments validate the effectiveness of our designs and demonstrate our framework’s strong performance in learning the universal and realistic policy.
 \end{itemize} 

\section{RELATED WORK}

\subsection{Door Manipulation Environment and Datasets}
Building a door manipulation environment that simulates the real world and transferring the policy trained in simulation to the real world has been the main approach for door manipulation tasks in recent years. 
However, recent works covering door manipulation have mainly two drawbacks: 
1)Unrealistic simulation. Besides using push or pull to open the door\cite{Mo_2021_ICCV, xu2022umpnet, wu2022vatmart, geng2023gapartnet, geng2023partmanip, eisner2022flowbot3d}, Robosuite~\cite{zhu2020robosuite} benchmarks opening doors with latching mechanism as standardized task but employs small doors with large handles. 
2)Lack of diversity in the dataset. 
PartNet-Mobility~\cite{xiang2020sapien} and  AKB-48~\cite{liu2022akb} provide a diverse dataset for articulated objects including doors. Focusing on the cross-category diversity, they ignore the intra-category diversity of doors. To address these two problems, we build a door manipulation environment with diverse realistic manipulation mechanisms, and equip this environment with a large-scale door dataset covering diverse types, handles and geometries for universal manipulation policy learning. 

\subsection{3D Articulated Object Manipulation}

Extensive investigations have been conducted in the field of articulated objects, encompassing various facets such as reconstructing the dynamic structure of articulated objects~\cite{abbatematteo2019learning, hausman2015active, katz2013interactive, nie2022structure, lv2022sagci, jiang2022ditto, du2023learning}, estimating 6d pose~\cite{jain2021screwnet, liu2020nothing, li2020category, weng2021captra, geng2023gapartnet}, comprehending the intricacies of manipulating them~\cite{katz2008manipulating, xu2022umpnet, eisner2022flowbot3d, geng2023partmanip, mu2021maniskill, gu2023maniskill2, geng2022end, borja2022affordance, schiavi2023learning}, and point-level visual affordance~\cite{Mo_2021_ICCV, wu2022vatmart, wang2021adaafford}. Among these areas of study, particular attention has been dedicated to the manipulation of articulated objects, with a specific emphasis on generating optimal action sequences.  Previous works~\cite{urakami2019doorgym, geng2023partmanip} have explored the training of a single universal policy using state-based ~\cite{urakami2019doorgym} or visual-based RL~\cite{geng2023partmanip, geng2022end}, which suffer from vast exploration space and complicated manipulation mechanisms for the door manipulation task. Otherwise, we propose a novel framework that disentangles the whole manipulation process into three stages with respective universal policies and integrates them into the whole universal policy leveraging conditioned training.

\begin{table*}[tb]
\begin{center}
\small
\setlength{\tabcolsep}{2mm}{
\begin{tabular}{c|ccc|ccc|ccc|ccc|ccc|ccc}

\hline
\multirow{2}{*}{\textbf{Datasets}} &\multicolumn{3}{c|}{\textbf{Int.}} & \multicolumn{3}{c|}{\textbf{Win.}} & \multicolumn{3}{c|}{\textbf{Car.}} & \multicolumn{3}{c|}{\textbf{Saf.}} & \multicolumn{3}{c|}{\textbf{Sto.}} & \multicolumn{3}{c}{\textbf{Ref.}} \\

 & \textbf{B} & \textbf{H} & \textbf{CO} & \textbf{B}  & \textbf{H} & \textbf{CO} & \textbf{B}  & \textbf{H}   & \textbf{CO} & \textbf{B} & \textbf{H}  & \textbf{CO} & \textbf{B}  & \textbf{H}   & \textbf{CO} & \textbf{B} & \textbf{H}  & \textbf{CO}\\ 
\hline
AKB-48~\cite{liu2022akb} & - & 9  & -  & -  & -  & -  & -   & -  & -  & -  & - & - & -   & -  & -  & -  & - & - \\
PartNet-Mobility~\cite{xiang2020sapien} & 26  & 22   & 26  & 3   & 1    & 3  & -  & -  & -  & 30 & 14   & 30  & 155   & -  & -  & 4  & - & - \\
GAPartNet~\cite{geng2023gapartnet} & 14  & 11   & 14  & -   & -    & -  & -  & -  & -  & 29 & 1   & 29  & 133   & -  & -  & 4  & - & - \\
DoorGym~\cite{urakami2019doorgym} & - & 20 & -   & -   & -   & -  & -   & -   & -  & - & -  & -  & -   & -  & -  & -  & - & - \\
\textbf{Ours} & \textbf{57} & \textbf{96} & \textbf{5472} & \textbf{18} & \textbf{37} & \textbf{666} & \textbf{22} & \textbf{15} & \textbf{330} & \textbf{61} & \textbf{39} & \textbf{2379} & \textbf{160}   & \textbf{8}  & \textbf{1280}  & \textbf{10}  & \textbf{9} & \textbf{90} \\
\hline
\end{tabular}
}

\caption{
   \textbf{Statistic comparisons} between previous dataset and ours. For category, \textbf{Int.}, \textbf{Win.}, \textbf{Car.}, \textbf{Saf.}, \textbf{Sto.}, \textbf{Ref.} respectively denote doors from Interior, Window, Car, Safe, StorageFurniture, Refrigerator. For asset number, \textbf{B}, \textbf{H}, \textbf{CO} indicate numbers of body, handle and composited object assets with the two parts. 
   }
\label{tab:datasets}
\end{center}
\vspace{-8mm}
\end{table*}
\begin{table}[tb]
\begin{center}
\small
\setlength{\tabcolsep}{2mm}{
\begin{tabular}{c|c|c|c|c|c}

\hline
\textbf{Env.} & \textbf{Data.} & \textbf{Mob.} & \textbf{Latch.} & \textbf{Part.} & \textbf{Occ.} \\
\hline
GAPartNet~\cite{geng2023gapartnet} & P + A  &    &   &   &   \\
W2A~\cite{Mo_2021_ICCV, wu2022vatmart, wang2021adaafford} & P &  &  & \includegraphics[width=0.04\linewidth]{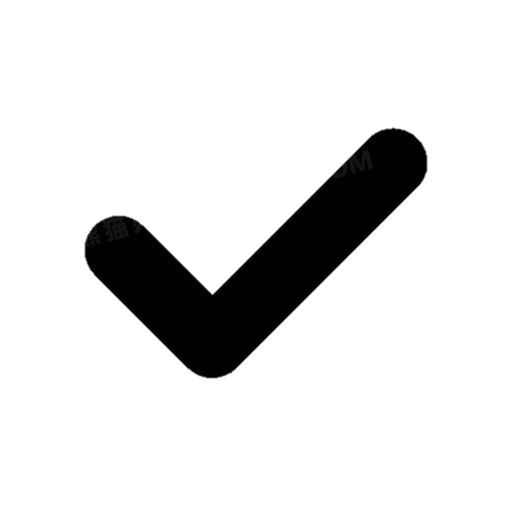}   &   \\ 
RLAfford~\cite{geng2022end} & P & \includegraphics[width=0.04\linewidth]{tabs/icons/hook1.jpg}  &   &   &  \\
PartManip~\cite{geng2023partmanip} & G & \includegraphics[width=0.04\linewidth]{tabs/icons/hook1.jpg} &   & \includegraphics[width=0.04\linewidth]{tabs/icons/hook1.jpg} &  \\
DoorGym~\cite{urakami2019doorgym} & D &  & \includegraphics[width=0.04\linewidth]{tabs/icons/hook1.jpg}   & \includegraphics[width=0.04\linewidth]{tabs/icons/hook1.jpg} &  \\
EnvAfford~\cite{wu2023learningenv} & P &   &   & \includegraphics[width=0.04\linewidth]{tabs/icons/hook1.jpg} & \includegraphics[width=0.04\linewidth]{tabs/icons/hook1.jpg}  \\
\textbf{Ours} & \textbf{Ours} & \includegraphics[width=0.04\linewidth]{tabs/icons/hook1.jpg} & \includegraphics[width=0.04\linewidth]{tabs/icons/hook1.jpg} & \includegraphics[width=0.04\linewidth]{tabs/icons/hook1.jpg} & \includegraphics[width=0.04\linewidth]{tabs/icons/hook1.jpg}\\
\hline
\end{tabular}
}
\caption{
   \textbf{Comparison between Our Environment and Others}. For simplicity, \textbf{Data.}, \textbf{Mob.}, \textbf{Latch.}, \textbf{Part.} and \textbf{Occ.} respectively denote Dataset, Mobile Robot Arm, Latching Mechanism, Partial Observation and Occlusion in Observation.
   Besides, \textbf{P}, \textbf{A}, \textbf{G} and \textbf{D} respectively denote PartNet-Mobility, AKB-48, GAPartNet and DoorGym in Table ~\ref{tab:datasets}.
   }
\label{tab:env}
\end{center}
\vspace{-8mm}
\end{table} 
\section{Large-scale Diverse Door Manipulation Environment}
An environment based on large-scale diverse door datasets and realistic simulation is a necessity for the training of a universal manipulation policy. Considering the lack of diversity and realness in the current simulation environments~\cite{urakami2019doorgym, zhu2020robosuite, Mo_2021_ICCV, xu2022umpnet, wu2022vatmart, geng2023gapartnet, geng2023partmanip, eisner2022flowbot3d}, we propose a novel environment with large-scale diverse door dataset (Section~\ref{env:dataset}) and realistic simulations (Section~\ref{env:simulation}) based on IsaacGym~\cite{makoviychuk2021isaac}.

\subsection{Large-scale Diverse Door Dataset}
\label{env:dataset}
In recent years, several works have proposed their datasets for door manipulation as illustrated in Table~\ref{tab:datasets}. DoorGym~\cite{urakami2019doorgym} claims a large-scale and scalable dataset specifically for door manipulation. Due to the hard-coded generation and the same templates, the doors they construct lack the diversity of geometry, which leads to bad performance on unseen doors. Limited by the environment, the door boards don't have a corresponding urdf or object entity, which can not be transferred to other environments. PartNet-Mobility~\cite{xiang2020sapien} and  AKB-48~\cite{liu2022akb} provide a diverse dataset for articulated objects including doors. Focusing on the cross-category diversity, they ignore the intra-category diversity of doors, which are not suitable for our goal. Hence we introduce a large-scale diverse dataset specifically for door manipulation.


Compared with the current dataset providing the whole door without part composition over diverse configuration, we construct our dataset in the formulation of two door parts, named \textit{body} and \textit{handle}. Given the composition method we provide, users of our dataset can configure their personalized door objects with diverse settings according to their needs. In total, we create 328 bodies and 204 handles which can be composed of thousands of door objects, as shown in Table~\ref{tab:datasets}.

To satisfy the need for diversity, we make efforts at both the category and intra-category level. For category level, We select 6 representative categories that cover most of the door-opening scenes we encounter in the real world, including \textbf{Interior}, \textbf{ Window}, \textbf{Car}, \textbf{Safe}, \textbf{Storagefurniture} and \textbf{Refrigerator}. 
For diversity in the intra-category, we collect hundreds of door bodies and handles with irregular geometries and diverse manipulation mechanisms as illustrated in Figure~\ref{fig:teasor}. Table~\ref{tab:datasets} provides a detailed statistic comparison between previous datasets and ours. Notice that we do the calculation for all object assets based on whether their manipulation mechanisms are similar to doors.


\subsection{Settings of Our Realistic Environment}
\label{env:simulation}
To closely emulate the real world, we specifically configure our environment both from the manipulation and the observation. To simulate the latching mechanism, we apply a force $\mathcal{F}_{door} \left(\theta \right)$ in the reverse direction of the opening to the doors until the handle joint angle $\theta_{h}$ have reached the opening threshold $thre_{door}$. Furthermore, we also add a resilient force to both the handle $\mathcal{F}_{handle} \left(\theta \right)$ and door to ensure the robustness of the manipulation policy. The forces are formulated as:
\begin{equation}
    \mathcal{F}_{door} \left(\theta_{d}, \theta_{h} \right) = \left\{
    \begin{alignedat}{2}
        F_f & , & \quad \theta_{h} & \leq thre \\
        k_1 \theta_{d} & , & \quad \theta_{h} & > thre \\
    \end{alignedat}
    \right.
\end{equation}
\begin{equation}
    \mathcal{F}_{handle} \left(\theta_{h} \right) = k_2 \theta_{h}
\end{equation}
Where $\theta_{d}$ represents the door joint angle. Here we set $k_1 = 3, k_2=3, F_f = 150$. The unit of force is $Newton$.
Instead of a parallel or suction gripper, we utilize the whole robot arm with a mobile base and parallel gripper as the agent to manipulate the doors. In this condition, collision between doors and robot arm as well as the joint limitation of the robot arm itself must be taken into consideration. Unlike previous works using the pseudo perfect observation~\cite{geng2022end, geng2023gapartnet, wu2022vatmart, Mo_2021_ICCV} or partial observation~\cite{geng2023partmanip, xu2022umpnet, eisner2022flowbot3d} which don't consider the occlusion caused by the door itself or the robot arm, we put a fixed camera in front of the door and robot arm and acquire the partially occluded point cloud generated by the depth image as the observation.
Table~\ref{tab:env} shows the detailed comparison between our environment and others from various aspects. 

\begin{figure*}[h]
    \centering
    \includegraphics[width=1\linewidth]{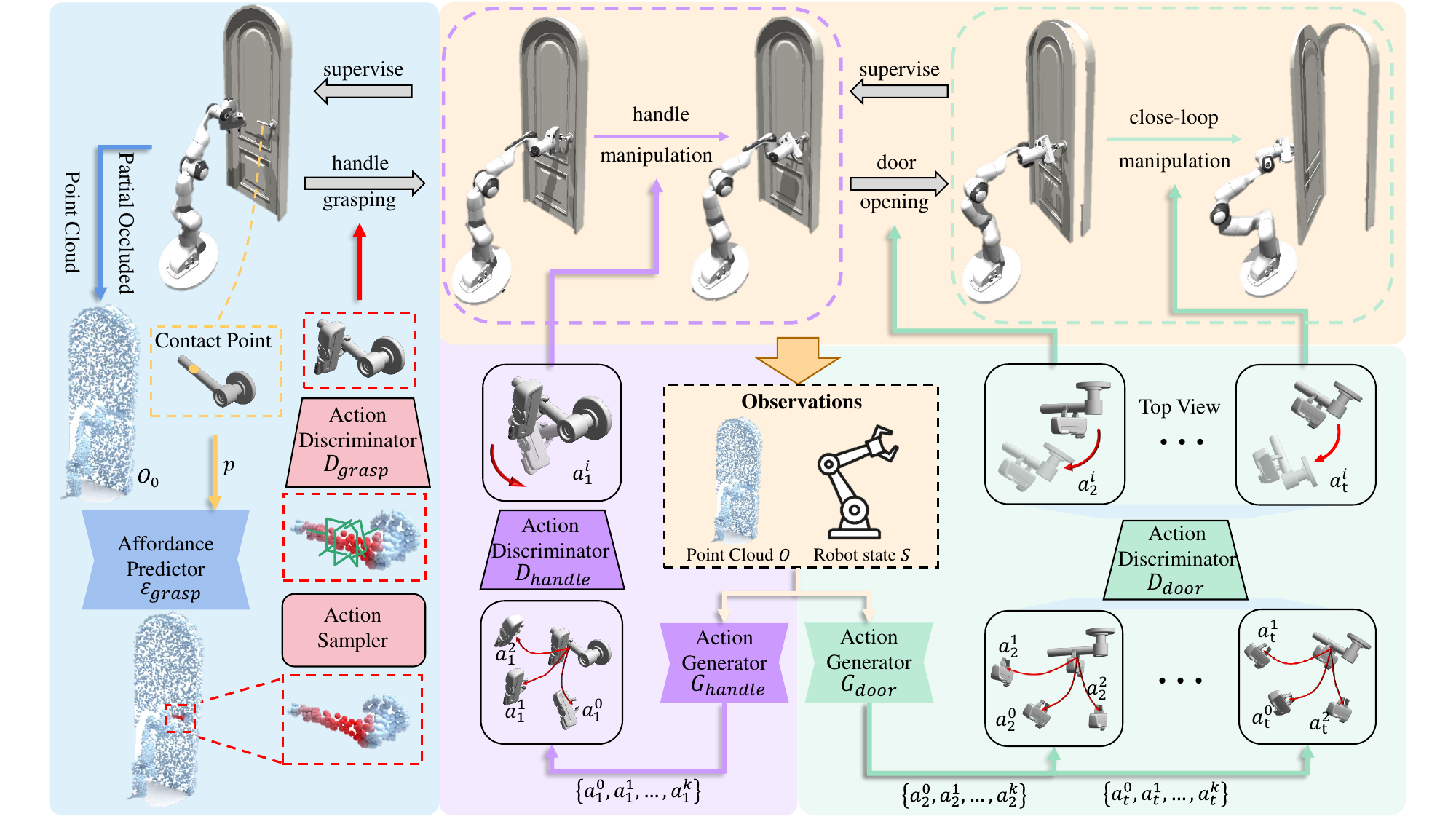}
    \caption{\textbf{Our pipeline for the framework}. We disentangle the entire door manipulation process into three stages: handle grasping, handle manipulation and door opening. We predict affordance map to graps the handle and employ the similar formulation but train separate policy for handle manipulation and door opening. Besides, We integrate the three policies for each stage leveraging conditioned training.}
    \label{fig:pipeline}
\vspace{-4mm}
\end{figure*}
\section{METHOD}
As illustrated in Figure~\ref{fig:pipeline}, we propose a novel framework that disentangles door manipulation into three distinct but related stages, each with a corresponding universal manipulation policy (Section~\ref{method:form}).
We leverage conditioned training to train these policies, as they have inter-dependencies, and thus they can be integrated into a unified universal policy  (Section~\ref{method:uni}). In the first stage, we employ generalizable point-level visual affordance~\cite{gibson1977theory, ning2023where2explore, zhao2022dualafford, ling2024articulated} to propose stable grasp poses (Section~\ref{method:handlegrasping}). 
In the second stage, we train a universal policy covering multiple handle manipulation mechanisms in our proposed realistic environment (Section~\ref{method:handlemanip}).
In the third stage, we train a policy to open doors with unlocked handles(Section~\ref{method:dooropening}). Additionally, we provide a comprehensive description of our data collection and training strategy(Section~\ref{method:data}).

\subsection{Problem Formulation}
\label{method:form}
We disentangle the door manipulation task into three stages:
handle grasping, handle manipulation and door opening.
Following UMPNet~\cite{xu2022umpnet}, we define the policy $\pi$ for each stage to be generating robot actions $a_t$, given the partial point cloud observation $O_t$ and the robot state $S_t$ at time $t$.
Here, the robot action $a_t=(p_t, r_t)$ indicates the next end-effector pose of robot arm, 
consisting of position $p_t \in \mathbb{R}^3$ and orientation $r_t \in \mathbb{SO}(3)$. 
The robot state $S_t$ consists of its joint angles and velocities.

In \textbf{stage one}, 
given the initial point cloud observation $O_0$ and the contact point $p_0 \in O_0$, the model outputs a per-point affordance map $A_{O_{0}}$ where a higher score on $p_0$ indicates a greater chance of successful door manipulation after grasping $p_0$ with action $a_0$. 

In \textbf{stage two}, given the observation $O_1$ after grasping, the goal is to manipulate the handle in the right direction to unlock the door. Given $O_1$ and robot state $S_1$, we train a manipulation policy to generate the action $a_1$ for diverse mechanisms.

In \textbf{stage three}, the goal is to open the door as wide as possible. Given the current visual observation $O_t$ and robot state $S_t$ at time $t$, we employ a manipulation policy to generate the opening action $a_t$ in a close-loop manner until the door joint angle $\theta_{d}$ exceeds a threshold.

\subsection{Disentanglement and Conditioned Training}
\label{method:uni}

As door manipulation includes multiple different stages,
it is difficult to directly train a universal policy in an end-to-end manner covering all the manipulation policies in those diverse stages.
Therefore, we disentangle the door manipulation into three distinct stages.
Due to the similarity of the manipulation policies in each stage, it is easy to train a universal and generalizable policy in each stage and merge them into a whole universal manipulation policy. 

While separated, these related stages have internal dependencies. For example, if the robot arm grasps the handle too far outward, it may result in disengagement while rotating the handle. Also, if the robot arm has a strange state in handle manipulation, it's easy to collide with the doors while the door is opening. Hence, to integrate all policies for each stage, we employ a conditioned training formulation leveraging these internal relations. Because we can evaluate the results of handle grasping during handle manipulation and evaluate the results of handle manipulation during door opening, we train the policy for door opening first. Then we train the handle manipulation policy using data collected based on the door opening policy. Given the policies for the latter two stages, we can obtain the first policy for initial handle grasping.

\subsection{Handle Grasping}
\label{method:handlegrasping}
Due to the diverse geometries of different handle types, it's challenging to train a universal and generalizable manipulation policy for initial handle grasping. While existing heuristic methods leveraging the pose estimation~\cite{geng2023gapartnet} or keypoint representation~\cite{wang2020learning} can not generalize to objects with irregular shapes, we employ point-level visual affordance~\cite{gibson1977theory} for manipulation, which predicts a point-level score map on the target object, indicating actionability for downstream tasks. 

The policy for handle grasping consists of three models: affordance predictor $\mathcal{E}_{\text {grasp }}$, action sampler and action discriminator $\mathcal{D}_{grasp}$. During inference, taking the initial observed point cloud $O_0 \in \mathbb{R}^{3 \times 4096}$ and the contact point $p \in \mathbb{R}^{3}$ as input, the affordance predictor $\mathcal{E}_{\text {grasp}}$  predicts a point-level score map where we choose point with highest score as the grasp point. We employ a segmentation version PointNet++~\cite{qi2017pointnet++} to extract a per-point feature from $O_0$, where $f_{p_0^i \mid O_0} \in \mathbb{R}^{128}$ represents the feature of point $p_0^i$, and use MLP to encode the contact point $p$ into a latent representation $f_{p_0} \in \mathbb{R}^{128}$. Based on the grasp point, the action sampler proposes alternative actions $\left\{a_0^0, a_0^1, \ldots,a_0^k\right\}$ by sampling random orientations from the normal plane of the handle axis and combining each of them with the grasp point. Taking the concatenation of the point cloud feature and the action feature extracted by another MLP, the action discriminator $\mathcal{D}_{grasp}$ outputs an action score. We select the action with the highest action score to grasp the handle. 

In the reverse order of inference, we train the model of action discriminator $\mathcal{D}_{grasp}$ firstly which is supervised by the final door joint angle $\theta_{d}$ using $L_1$ loss. For training affordance predictor $\mathcal{E}_{\text {grasp }}$, we sample 50 random actions based on the contact point $p_0$ selected from $O_0$. Then we estimate their action scores using $\mathcal{D}_{grasp}$ and regress the prediction to the mean score of the top-10 rated action with $L_1$ loss. 

\subsection{Handle Manipulation}
\label{method:handlemanip}
Leveraging these diverse manipulation mechanisms provided by our proposed environment, we employ a generative network architecture to learn the high dimensional data distribution and output the handle manipulation action. 
The policy for handle manipulation is composed of two modules: action generator $\mathcal{G}_{handle}$ and action discriminator $\mathcal{D}_{handle}$. 
The action generator $\mathcal{G}_{handle}$ is implemented as a conditional variational autoencoder(cVAE)~\cite{sohn2015learning}, composed of an action encoder that maps the input action $a_{t}$ into a Gaussian noise $z$ and an action decoder that reconstructs the action input from the noise vector. We set the current point cloud $O_t$ and robot state $S_t$ as the conditions of the encoder and decoder, which are implemented both by MLPs.
Given the alternative actions $\left\{a_1^0, a_1^1, \ldots,a_1^k\right\}$ generated by $\mathcal{G}_{handle}$, plus $O_t$ and $S_t$, the action discriminator predict action scores which we select the action with the highest score as the manipulation action.

For training action generator $\mathcal{G}_{handle}$, we use the KL divergence loss for regularizing Gaussian bottleneck noises, the $L_1$ loss to regress the action position $p_t$ and the a 6D-rotation loss~\cite{zhou2019continuity} for supervise the action orientation$r_t$. The total loss can be formulated as:
\begin{equation}
    \mathcal{L}_{h} = \lambda_{KL}\mathcal{L}_{kl} + \lambda_{pos}\mathcal{L}_{pos} + \lambda_{rot}\mathcal{L}_{rot}
\end{equation}
where we use a set of hyper-parameters to adjust the loss balance.
For training action discriminator $\mathcal{D}_{handle}$, we use $L_1$ loss to supervise the action score with the residual handle joint angle between before and after the action execution.

\subsection{Door Opening}
\label{method:dooropening}
To tackle the diverse geometries and long action sequences, 
Instead of open-loop manipulation~\cite{wu2022vatmart}, we employ a closed-loop formulation that iteratively generates the next action $a_t$ for door opening based on the current observation including the partially occluded point cloud $O_t$ and the robot state $S_t$ for door opening stage. We follow the implementation of the policy for handle manipulation and employ an action generator $G_{door}$ and an action discriminator $D_{door}$ to output the door opening action. Different from the previous discriminator, we supervise the action score with the residual door joint angle between before and after the action execution.

\begin{figure*}
    \centering
    \includegraphics[width=1 \linewidth]{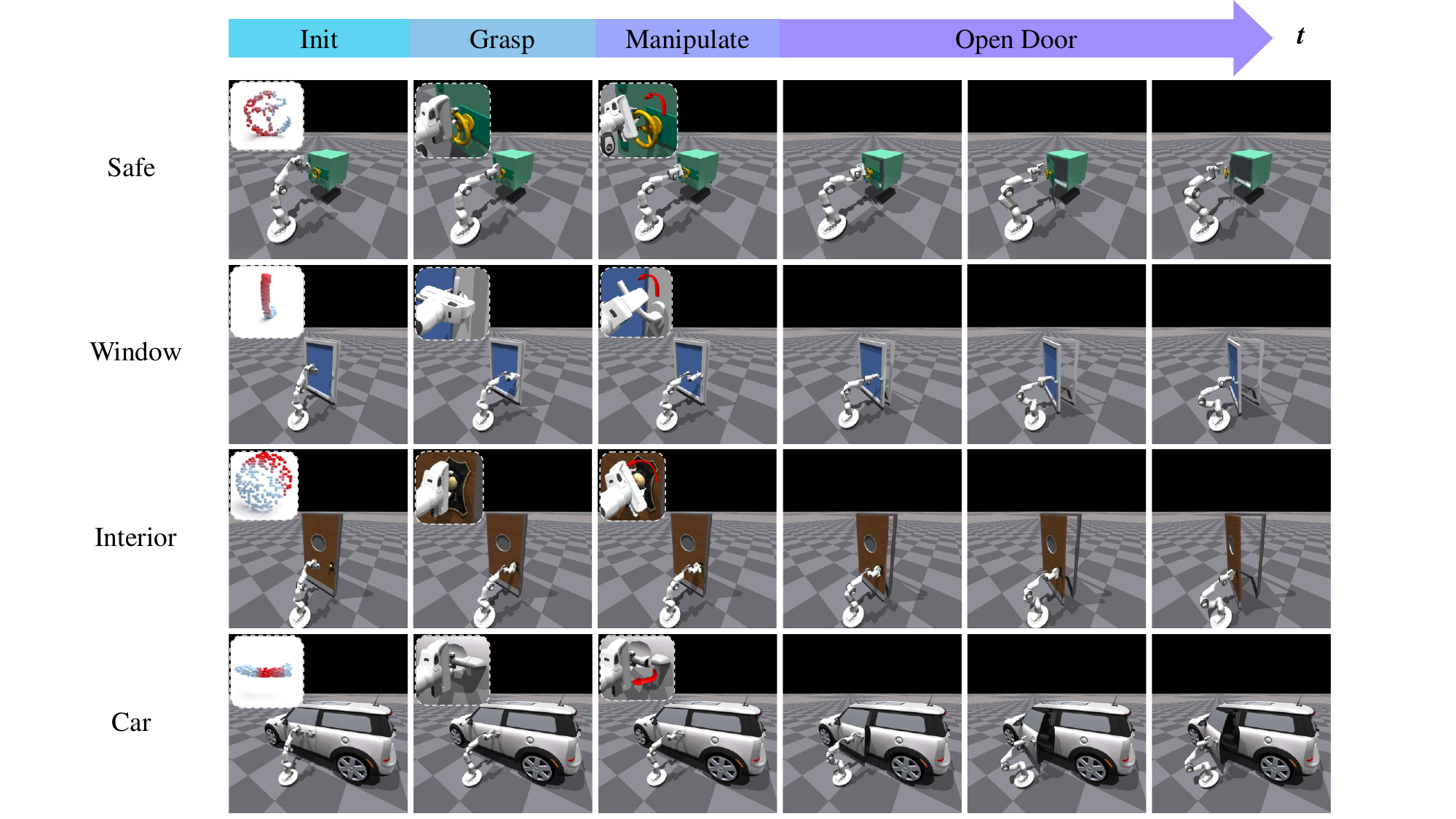}
    \caption{\textbf{Manipulation Sequence Guided by Our Universal Manipulation Policy.}} 
    
    \label{fig:exp}
\vspace{-4mm}
\end{figure*}
\subsection{Data Collection and Training Strategy}
To train policies for the three stages, we collect data including the input observation and ground truth supervision leveraging the rule-based method. For different manipulation mechanisms and stages, we use the door states like the joint axis of the handle and body which can be acquired in the simulation environment to calculate the next action. 

In the reverse order of inference, we train the door-opening policy first. By supervising the policy with the final door manipulation result, the door opening policy can generate action that facilitates to open the door as wide as it can. Then we replace the rule for the door opening with the trained policy and collect data for handle manipulation policy training. In this way, the handle manipulation policy is integrated with the door opening policy, which outputs action that facilitates door opening. Finally, we replace the latter two rules with trained two policies and collect data for affordance training. Due to the supervision of the final door joint angle for affordance score, the policy for the initial handle grasping can predict a foresightful grasp pose which helps to avoid the bad cases in the following stages, like the disengagement between the end-effector and handle, the collision between the robot arm and door. 

\label{method:data}

\vspace{-2mm}
\section{EXPERIMENTS}

\subsection{Task, Metric, Evaluation and Settings}
We conduct our experiments on the representative door manipulation tasks: \textbf{pull door}. The initial closed door can only be unlocked and pulled to open. The robot arm needs to pull the door until the door joint angle $\theta_{d}$ is larger than a threshold $thre_{door}$. Here, we set $thre_{door}$ to be $45^\circ$. 

For \textbf{training and testing}, to show the generalizability of our proposed framework across diverse categories, geometries and manipulation mechanisms, we train the universal manipulation policy on the door category including Interior, Window, Safe and Car. We carefully split the object assets of the aforementioned categories into the train and test part which ensures the universal policy is tested on unseen shapes. Furthermore, we also leave two more categories, Storagefurniture and Refrigerator, for extensive tests on unseen novel categories. 

For \textbf{evaluation metric}, we use the task success rate as our main evaluation metric. To reduce the evaluation noise, we conduct each experiment 3 times using random seeds and report the mean performance as well as the variance. 

\subsection{Baselines}
We compare our method with the following baselines: 1) \textbf{GAPartNet + GT}: following the heuristic manipulation method of GAPartNet, we directly obtain the initial ground truth like poses of the handle and board from simulation and implement a heuristic motion planning method for door manipulation. 2)\textbf{DoorGym}: we introduce the state-based PPO~\cite{schulman2017proximal} implemented in DoorGym as our baseline. Here we use a flying gripper as the agent and put the gripper close to the handle in the beginning. 3)\textbf{PartManip}: we introduce a visual-based PPO used in PartManip as our baseline. The setting is similar to \textbf{DoorGym}. 4)\textbf{VAT-MART}: a method that proposes visual action trajectory for downstream manipulation task in an open-loop formulation. For a fair comparison, We develop an implementation version in our environment.

\subsection{Result Analysis}
Figure~\ref{fig:exp} shows the whole manipulation sequence of our universal manipulation in four categories, including Safe, Window, Interior and Car. The results show that our universal policy can generalize over diverse categories, geometries and manipulation mechanisms. Located in the upper left of the initial pictures, we use a heap map to visualize the affordance score where the redder point indicates the higher score. It's worth mentioning that the affordance map represents where the fingers of the end-effector contact with the object, which just matches the result affordance of the round handle for the category Interior.

Table ~\ref{tab:task} shows that our framework outperforms all baselines for both the train and test  categories in the pull door task. When encountering handles with irregular types and geometries, \textbf{GAPartNet + GT} can not calculate a proper initial grasp pose even provided with ground truth handle pose. Besides, the heuristic motion planning method can not generalize well in unseen objects. For the two RL-based methods \textbf{DoorGym} and \textbf{PartManip}, due to the large exploration space for multiple stages and complicated reward engineering for diverse categories, these methods fail to train an end-to-end policy for door manipulation. Open-loop method \textbf{VAT-MART} predicts the action trajectory with the initial observation, which leads to accumulation error and inability to real-time adjustments.

\begin{table}[tb]
\begin{center}
\small
\setlength{\tabcolsep}{1.5mm}{
\begin{tabular}{c|cccc|cc}

\hline
\textbf{Task} & \multicolumn{6}{c}{\textbf{Pull Door}}\\

\hline
\multirow{2}{*}{\textbf{Method}}
 & \multicolumn{4}{c|}{\textbf{Train}}
 & \multicolumn{2}{c}{\textbf{Test}}\\

 & \includegraphics[width=0.04\linewidth]{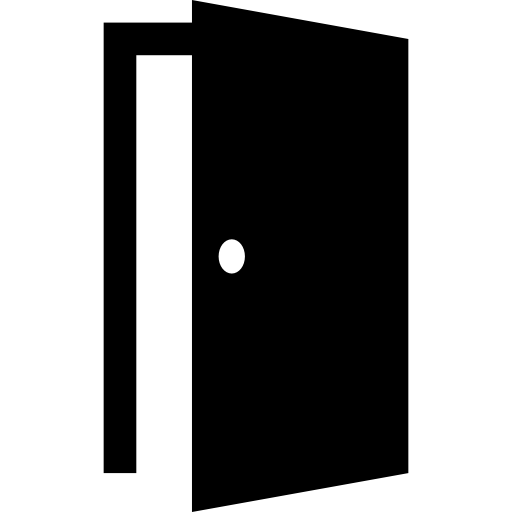}
 & \includegraphics[width=0.04\linewidth]{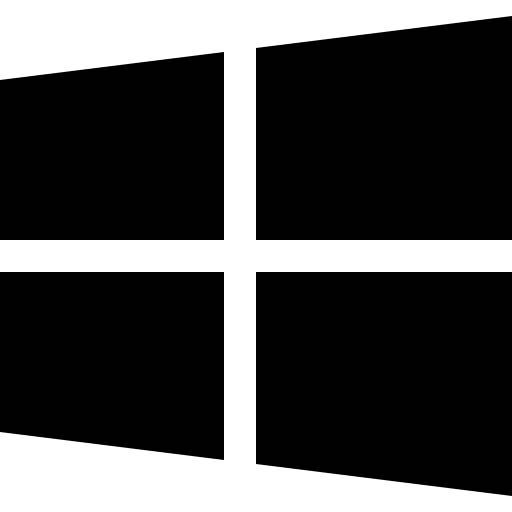} 
 & \includegraphics[width=0.08\linewidth]{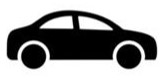} 
 & \includegraphics[width=0.04\linewidth]{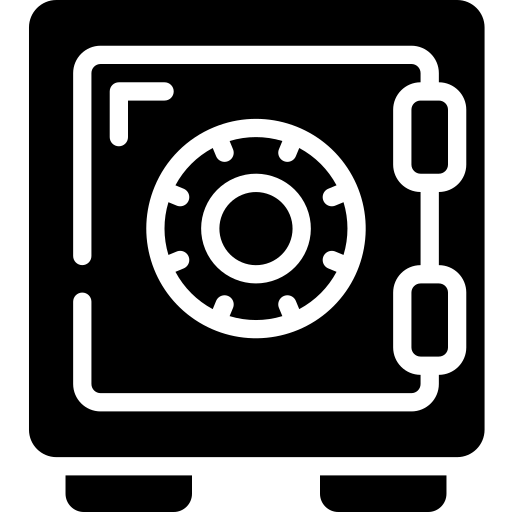} 
 & \includegraphics[width=0.04\linewidth]{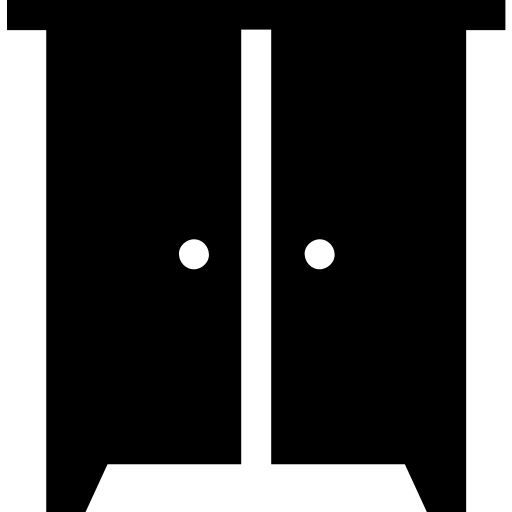}
 & \includegraphics[width=0.04\linewidth]{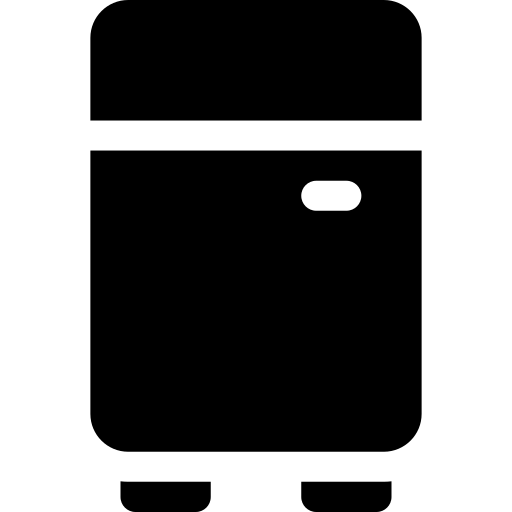}
 \\
 
\hline 
 GAPartNet~\cite{geng2023gapartnet}+GT & 0.62 & 0.88 & 0.41 & 0.44 & 0.52 & 0.26\\
 DoorGym~\cite{urakami2019doorgym} & 0.56 &  0.72 & 0.61  &  0.41 & 0.19 &  0.23 \\
 PartManip~\cite{geng2023partmanip} & 0.47 & 0.61  &  0.54 & 0.34  & 0.42  & 0.19\\
 VAT-MART~\cite{wu2022vatmart} & 0.59  &  0.62  & 0.57  &  0.43  & 0.51  & 0.25 \\
\hline
 Ours w/o disentangle. & 0.44 & 0.88  & 0.20  & 0.19  & 0.05  & 0.22\\
 Ours w/o condition. & 0.77 & 0.31  & 0.58  & 0.51  & 0.54  & 0.33\\
 Ours w/o state. & 0.73 & 0.59  & 0.16  & 0.36  & 0.45  & 0.37\\
 Ours w/o mobile. & 0.87 & 0.60  & 0.00  & 0.43  & 0.50  & 0.81\\
 \textbf{Ours} &\textbf{0.99}  & \textbf{0.91}  & \textbf{0.81}  & \textbf{0.72}  & \textbf{0.75}  & \textbf{0.89}\\
\hline
\end{tabular}
}

\caption{
   \textbf{Experimental results of the baselines and ablation studies on Pull Door task. } In the \textbf{Method} column, \textbf{Train} represents we test on unseen shapes in train categories. \textbf{Test} represents we test on unseen shapes in unseen test categories.
   }
\label{tab:task}
\end{center}
\vspace{-6mm}
\end{table}
\begin{figure}
    \centering
    \includegraphics[width=0.9 \linewidth]{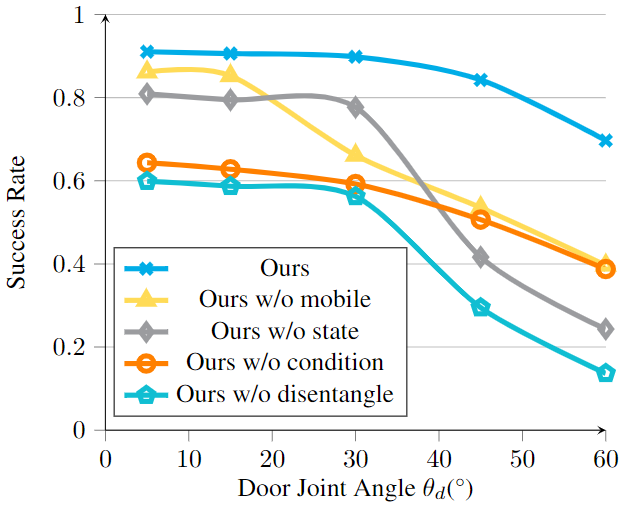}
    \caption{
    \textbf{Comparison of the ablations and ours for different door joint angles}. Here for each door joint angle, we do experiments on all categories and get the average success rate. }
    \label{fig:line}
\vspace{-3mm}
\end{figure}
\begin{figure}
    \centering
    \includegraphics[width=1\linewidth]{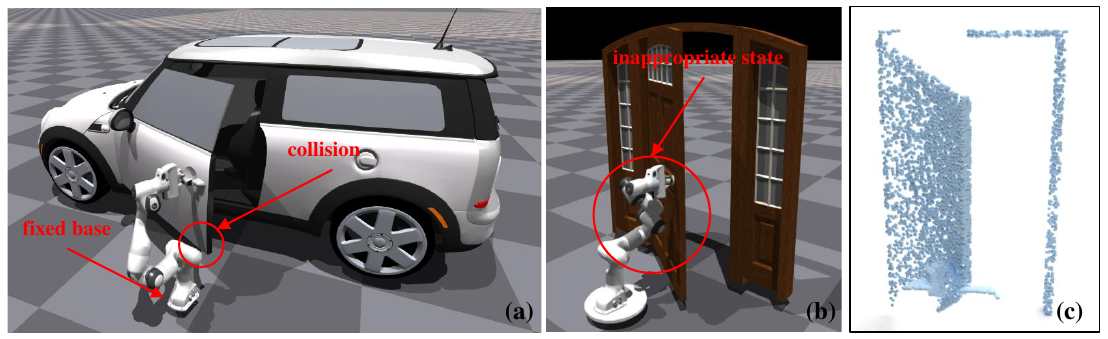}
    \caption{\textbf{Failure cases of ablated versions.}}
    \label{fig:fail}
\vspace{-4mm}
\end{figure}

\subsection{Ablation Study}
To further evaluate the importance of different components of our method, we conducted an ablation study by comparing our method with four ablations: 1) \textbf{Ours w/o disentangle.}: instead of training policies for the latter two stages separately, we use a single policy to generate the manipulation action after handle grasping. 2)\textbf{Ours w/o condition.}: ours without conditioned training. 3)\textbf{Ours w/o state.}: ours without the input observation of robot state. 4)\textbf{Ours w/o mob.}: ours that uses a fixed base robot arm instead of a mobile robot arm.


The experimental results presented in Table~\ref{tab:task} and Figure~\ref{fig:line} provide evidence supporting the efficacy of the four components we have devised in enhancing the performance of the universal manipulation policy. Figure~\ref{fig:line} depicts the trends observed in the performance of \textbf{Ours} and \textbf{Ours w/o condition.}  as the door joint angle $\theta_{d}$ increases. Both approaches exhibit a similar downward trend, indicating that conditioned training contributes to performance improvement. Conversely, \textbf{Ours w/o disentangle.} yields the lowest success rate across most categories and all $\theta_d$ values due to the inappropriate state(Figure~\ref{fig:fail}(b)), underscoring the criticality of disentangling the door manipulation process for effective universal manipulation policy learning. Notably, \textbf{Ours w/o mobile.} experiences a rapid decline as $\theta_d$ slightly increases and achieves a zero success rate in the Car category, suggesting that the absence of a mobile base makes it prone to collisions with doors as shown in Figure~\ref{fig:fail} (a). Finally, \textbf{Ours w/o state.} exhibits a rapid decline in performance after $\theta_d$ reaches $30^\circ$, demonstrating that incorporating robot state information can mitigate degradation in point cloud data(Figure~\ref{fig:fail}(c)) as $\theta_d$ increases.

\begin{figure}
    \centering
    \includegraphics[width=0.95\linewidth]{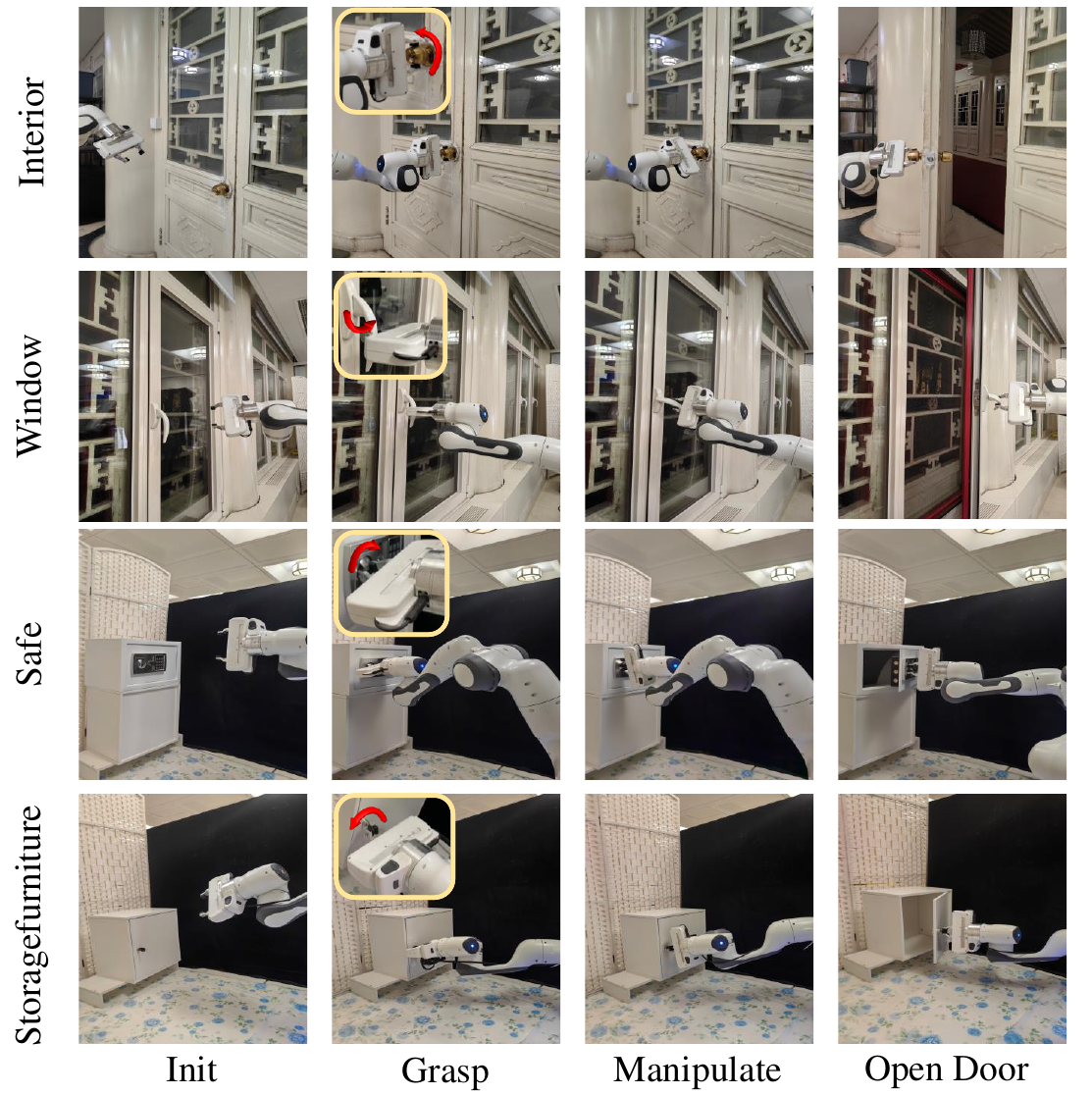}
    \caption{\textbf{Real-World Experiments.}}
    \label{fig:real_exp}
\vspace{-4mm}
\end{figure}
\subsection{Real-world Experiment}
We conduct experiments that involve interacting with various real-world doors including Interior, Window, Safe and Storagefurniture as shown in Figure~\ref{fig:real_exp}. We employ a Franka Emika Panda Robot Arm to perform door manipulation tasks. For input of observation, we obtain real-time point clouds from an Azure Kinect DK depth camera and robot state from the robot arm. The manipulation process in Figure~\ref{fig:real_exp} demonstrates that our universal policy can effectively transfer to real-world scenarios. See supplementary for more details.

\section{CONCLUSION}
In this work, we aim to learn a universal manipulation policy for doors with diverse categories, geometries and mechanisms. For realistic simulation, we introduce a novel door manipulation environment including a large-scale door dataset and diverse realistic latching mechanisms. Based on the environment, we present a novel framework for universal policy learning which disentangles the entire manipulation process into three stages and integrates them by reverse training. Extensive experiments validate the effectiveness of our designs and demonstrate our framework’s stronger performance than previous work. 

{
    \small
    \bibliographystyle{ieeenat_fullname}
    \bibliography{main}
}


\end{document}